%% file: camera-ready-srl4orl.tex
\documentclass[11pt,a4paper]{article}
\usepackage[hyperref]{naaclhlt2018}
\usepackage{times}
\usepackage{latexsym}
\usepackage{url}

\usepackage{lingmacros}
\usepackage{booktabs}
\usepackage{xcolor}
\usepackage{tabularx,pbox}
\usepackage{amsmath}
\usepackage{mathtools}
\DeclarePairedDelimiter\ceil{\lceil}{\rceil}

\definecolor{bothcolor}{RGB}{172,229,163}
\definecolor{fscolor}{RGB}{168,211,238}
\definecolor{stlcolor}{RGB}{254,255,184}

\usepackage{enumerate}

\aclfinalcopy 
\def\aclpaperid{423} 

\title{SRL4ORL: Improving Opinion Role Labeling Using\\ Multi-Task Learning With Semantic Role Labeling}

\author{Ana Marasovi\'c \and Anette Frank\\ 
  Research Training Group AIPHES \\
  Department of Computational Linguistics\\
  Heidelberg University \\
  {\tt \{marasovic,frank\}@cl.uni-heidelberg.de} }
\date{}

\begin{document}
\maketitle
\begin{abstract}
For over a decade, machine learning has been used to extract {\em opinion-holder-target structures} from text to answer the question {\em Who expressed what kind of sentiment towards what?}. Recent neural approaches do not outperform the state-of-the-art feature-based models for Opinion Role Labeling (ORL). We suspect this is due to the scarcity of labeled training data and address this issue using different
multi-task learning (MTL) techniques with a related task which has substantially more data, i.e.\ Semantic Role Labeling (SRL). We show that two MTL models improve significantly over the single-task model for labeling of both holders and targets, on the development and the test sets. We found that the vanilla MTL model which makes predictions using only shared ORL and SRL features, performs the best. With deeper analysis we determine what works and what might be done to make further improvements for ORL. 
\end{abstract}

\input{intro}

\input{models}
\input{setup}
\input{results}

\input{analysis}
\input{related}

\input{conclusions}

\section*{Acknowledgments} This work has been supported by the German Research Foundation as part of the Research Training Group Adaptive Preparation of Information from Heterogeneous Sources (AIPHES) under grant No.\ GRK 1994/1.
%

\bibliography{references}
\bibliographystyle{acl_natbib}

\appendix
\label{sec:supplemental}
\input{supplement}

\end{document}

%% file: intro.tex
\section{Introduction}
\label{sec:intro}

Fine-Grained Opinion Analysis (FGOA) aims to: 
(i) detect opinion expressions (O) that convey attitudes such as sentiments, agreements, beliefs or intentions (like \textit{feared} in example (\ex{1})), (ii) measure their intensity (e.g.\ strong), (iii) identify their holders (H), i.e.\ entities that express an attitude (e.g.\ \textit{it}), (iv) identify their targets (T), i.e.\ entities or propositions at which the attitude is directed (e.g.\ \textit{violence}) and (v) classify their target-dependent attitude (e.g.\ negative sentiment)\footnote{Examples are drawn from MPQA \citep{Wiebe2005AnnotatingEO}.}.

\enumsentence{Australia said [it]$_{H}$ [\textbf{feared}]$_{O_{neg}}$ [violence]$_{T}$ if voters thought the election had been stolen.}

As the commonly accepted benchmark corpus MPQA \citep{Wiebe2005AnnotatingEO} uses span-based annotations to represent \textit{opinion entities} (opinions, holders and targets), the task is usually approached with sequence labeling techniques and the BIO encoding scheme \citep{choi2006joint,Yang2013JointIF, Katiyar2016InvestigatingLF}. Initially pipeline models were proposed which first predict opinion expressions and then, given an opinion, label its \textit{opinion roles}, i.e.\ holders and targets \citep{Kim2006ExtractingOO, Johansson2013RelationalFI}. Pipeline models have been substituted with so-called joint models that simultaneously identify all opinion entities, and predict which opinion role is related to which opinion \citep{choi2006joint,Yang2013JointIF, Katiyar2016InvestigatingLF}. Recently an LSTM-based joint model was proposed \citep{Katiyar2016InvestigatingLF} that unlike the prior work \citep{choi2006joint,Yang2013JointIF} does not depend on external resources (such as syntactic parsers or named entity recognizers). The neural variant does not outperform the feature-based CRF model \citep{Yang2013JointIF} in Opinion Role Labeling (ORL).

\input{tables/srl_demo}

Both the neural and the CRF joint models achieve about 55\% F1 score for predicting which targets relate to which opinions in MPQA. Thus, these models are not yet ready to answer the question this line of research is usually motivated with: \textit{Who expressed what kind of sentiment towards what?}. Our goal is to investigate the limitations of neural models in solving different subtasks of FGOA on MPQA and to gain a better understanding of what is solved and what is next.

We suspect that one of the fundamental obstacles for neural models trained on MPQA is its small size. One way to address scarcity of labeled data is to use multi-task learning (MTL) with appropriate auxiliary tasks. A promising auxiliary task candidate for ORL is Semantic Role Labeling (SRL), the task of predicting predicate-argument structure of a sentence, which answers the question \textit{Who did what to whom, where and when?}. Table \ref{tbl:srl_demo} illustrates the output of the SRL demo\footnote{\url{http://barbar.cs.lth.se:8081}} for example (\ex{0}), following the PropBank SRL scheme \cite{Palmer2005ThePB}\footnote{Henceforth we use the PropBank SRL framework.}.

\textbf{SRL4ORL.} The semantic roles of the predicate \textit{fear} (marked blue bold) correspond to the opinion roles H and T, according to MPQA. For this reason, the output of SRL systems has been commonly used for feature-based FGOA models \citep{Kim2006ExtractingOO, Johansson2013RelationalFI, choi2006joint,Yang2013JointIF}. Additionally, a considerable amount of training data is available for training SRL models (Table \ref{tbl:data_stats_1} in Sec.\ \ref{sec:setup}), which made neural SRL models successful \citep{zhou-xu:2015:ACL-IJCNLP,yang-mitchell:2017:EMNLP2017}. 

\textbf{Obstacles.} Although SRL is similar in nature to ORL, it cannot solve ORL for all cases \cite{Ruppenhofer2008FindingTS}. In example (\ex{1}) holder and target of the predicate \textit{please} correspond to A1, A0 semantic roles respectively, wheres for the predicate \textit{fear} in (\ex{0}) holder and target correspond to A0, A1 respectively. We took into account this observation when deciding on an appropriate MTL model by splitting its parameters into shared and task-specific ones (i.e.\ hard-parameter sharing).

\enumsentence{[I]$_{H}^{A1}$ am very [\textbf{pleased}]$_{O_{pos}}$ that [the Council has now approved the Kyoto Protocol thus enabling the EU to proceed with its ratification]$_{T}^{A0}$.}

A further obstacle for properly exploiting SRL training data with MTL could be specificities, inconsistency and incompleteness of the MPQA annotations. In example (\ex{1}), \textit{Rice} expressed his negative sentiment towards \textit{the three countries in question} by \textit{setting the criteria} which states something negative about those countries: they \textit{are repressive and grave human rights violators [...]}. In this case, the model should not pick any local semantic role for the target.

\enumsentence{The criteria [\textbf{set by}]$_{O_{neg}}$ [Rice]$_{H}$ are the following: [the three countries in question]$_{T}$ are repressive and grave human rights violators, and aggressively seeking weapons [...].}

In examples (\ex{1}--\ex{2}), the same opinion expression \textit{concerned} realizes different scopes for the target. A model which exploits SRL knowledge could be biased to always label targets as complete SRL role constituents, as in example (\ex{2}).

\enumsentence{Rice told us [the administration]$_{H}$ was [\textbf{concerned}]$_{O_{neg}}$ that [Iraq]$_{T}$ would take advantage of the 9/11 attacks.}
\enumsentence{[The Chinese government]$_{H}$ is deeply  [\textbf{concerned}]$_{O_{neg}}$ about [the sudden deterioration in the Middle East situation]$_{T}$, Tang said.} 

Regarding incompleteness, prior work \cite{Katiyar2016InvestigatingLF} has shown that their model makes reasonable predictions in sentences which do not have annotations at all, e.g.\ [mothers]$_{H}$ [\textbf{care}]$_{O}$ for [their young]$_{T}$, in: \textit{From the fact that mothers care for their young, we can not deduce that they ought to do so, Hume argued.} 

The examples above show that incorporating SRL knowledge via multi-task learning is a reasonable way to improve ORL, but at the same time they alert us that given the specificities of MPQA and ORL annotations in general, it is not obvious whether MTL can overcome divergences in the annotation schemes of opinion and semantic role labeling. We investigate this research question by adopting one of the recent successful architectures for SRL \cite{zhou-xu:2015:ACL-IJCNLP} and experiment with different multi-task learning frameworks. 

Our contributions are: (i) we adapt a recently proposed neural SRL model for ORL, (ii) we enhance the model using different MTL techniques with SRL to tackle the problem of scarcity of labeled data for ORL, (iii) we show that most of the MTL models improve the single-task model for labeling of both holders and targets on development and test sets, and two of them make yield significant improvements, (iv) by deeper analysis we provide a better understanding of what is solved and where to head next for neural ORL.

%% file: tables/srl_demo.tex
\begin{table*}[t]
\centering
\resizebox{\textwidth}{!}{
\centering
\begin{tabular}{@{}c@{~~}c@{~~}c@{~~}c@{~~}c@{~~}c@{~~}c@{~~}c@{~~}c@{~~}c@{~~}c@{~~}c@{~~}c@{~~}c@{~~}c@{}}
   & Australia & said & it & feared & violence & if     & voters & thought & the    & election & had    & been   & stolen & . \\
say.01   & A0        & -    & A1 & A1     & A1       & A1     & A1     & A1      & A1     & A1       & A1     & A1     & A1     & - \\
fear.01  & -         & -    & \textcolor{blue}{\textbf{A0}} & -      & \textcolor{blue}{\textbf{A1}}       & AM-ADV & AM-ADV & AM-ADV  & AM-ADV & AM-ADV   & AM-ADV & AM-ADV & AM-ADV & - \\
think.01 & -         & -    & -  & -      & -        & -      & A0     & -       & A1     & A1       & A1     & A1     & A1     & - \\
steal.01 & -         & -    & -  & -      & -        & -      & -      & -       & A1     & A1       & -      & -      & -      & -
\end{tabular}
}
\caption{Output of the SRL demo.}
\label{tbl:srl_demo}
\end{table*}

%% file: models.tex
\section{Neural MTL for SRL and ORL} 
\label{sec:models}

Neural multi-task learning (MTL) receives a lot of attention and new MTL architectures emerge regularly. Yet there is no clear consensus which MTL architecture to use in which conditions. We experiment with well-received architectures that could adapt to different cases of ORL from Section \ref{sec:intro}.

\begin{figure}[t]
\minipage[t]{0.22\textwidth}
  \centering
  \includegraphics[width=0.75\linewidth]{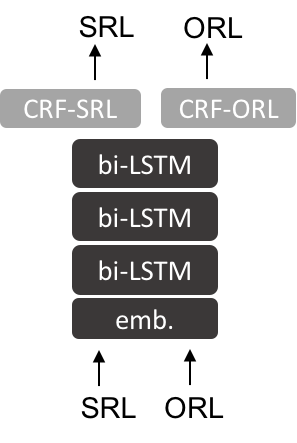}
  \caption{Fully-sha\-red (FS) MTL.}\label{fig:fs}
\endminipage\hfill
\minipage[t]{0.22\textwidth}%
  \centering
  \includegraphics[width=0.8\linewidth]{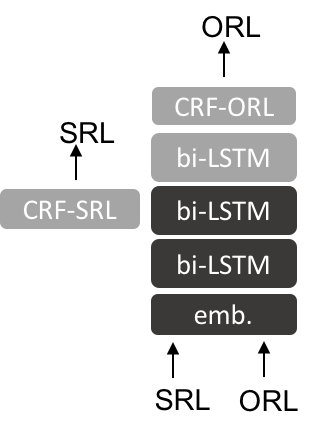}
 \caption{Hier\-ar\-chi\-cal\- MTL (H-MTL). }\label{fig:hmtl}
\endminipage
\end{figure}

As a general neural architecture for single- and multi-task learning we use the recently proposed SRL model \cite{zhou-xu:2015:ACL-IJCNLP} (Z\&X-STL) which successfully labels semantic roles without any syntactic guidance. This model consists of a stack of bi-directional LSTMs and a CRF which makes the final prediction. The inputs to the first LSTM are not only token embeddings but three additional features: embedding of the predicate, embedding of the context of the predicate and an indicator feature (1 if the current token is in the predicate context, 0 otherwise). Thus, every sentence is processed as many times as there are predicates in it. Adapting this model for labeling of opinion roles is straightforward, the only difference being that opinion expressions can be multi-words and only two opinion roles are assigned.    

MTL techniques aim to learn several tasks jointly by leveraging knowledge from all tasks. In the context of neural networks, MTL is commonly used in such a way that it is predefined which layers have tied parameters and which are task-specific (i.e.\ hard-parameter sharing). There are various ways of defining which parameters should be shared and how to train them. 

\textbf{Fully-shared (FS) MTL model.} A fully-shared model (Fig.\ \ref{fig:fs}) shares all parameters of the general model except the output layer. Each task has a task-specific output layer which makes the prediction based on the representation produced by the final LSTM. When training on a mini-batch of a certain task, parameters of the output layer of the other tasks are not updated. This model should be effective for constructions with a clear mapping between opinion and semantic roles such as \{H $\mapsto$ A0, T $\mapsto$ A1\} as in example (1) (Sec.\ \ref{sec:intro}).  

\textbf{Hierarchical MTL (H-MTL) model.} For NLP applications, often some given (high-level) task is supposed to benefit from another (low-level) task more than the other way around, e.g.\ parsing from POS tagging. This intuition lead to designing hierarchical MTL models \cite{sogaard2016deep, hashimoto2016joint} in which predictions for low-level tasks are not made on the basis of the representation produced at the final LSTM, but on the representation produced by a lower-layer LSTM (Fig.\ \ref{fig:hmtl}). Task-specific layers atop shared layers could potentially give the model more power to distinguish or ignore certain semantic roles. If so, this MTL model is more suitable for examples like (2) and (3) (Sec.\ \ref{sec:intro}). 

\begin{figure}[t]
  \centering
  \includegraphics[width=0.6\linewidth]{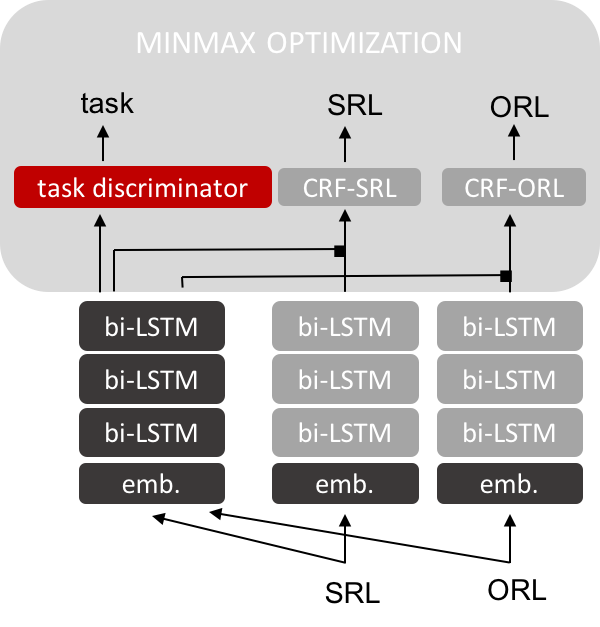}
  \caption{(Adversarial) state-private ((A)SP) MTL.}\label{fig:asp}
\end{figure}

\textbf{Shared-private (SP) MTL model.} In the state-private model, in addition to the stack of shared LSTMs, each task has a stack of task-specific LSTMs \cite{liu2017adversarial} (Fig.\ \ref{fig:asp}). Representations at the outermost shared LSTM and the task-specific LSTM are concatenated and passed to the task-specific output layer. The ORL representation produced independently from SRL gives the model the ability to utilize the shared and entirely task-specific information. For labeling of targets, it is expected that for examples (1) \& (5) the model relies mostly on the shared representation, for examples (2) \& (4) on both shared and ORL-specific representations, and for example (3) solely on the ORL-specific representation.

\textbf{Adversarial shared-private (ASP) model.} The limitation of the SP model is that it does not prevent the shared layers from capturing task-specific features. To ensure this, ASP extends the SP model with a \textit{task discriminator} \cite{liu2017adversarial}. The task discriminator (Fig.\ \ref{fig:asp}, marked red) predicts to which task the current batch of data belongs, based on the representation produced by the shared LSTMs. If the shared LSTMs are task-invariant, the discriminator should perform badly. Thus, we update the shared parameters to maximize the discriminator's cross-entropy loss. At the same time we want the discriminator to challenge the shared LSTMs, so we update the discriminator's parameters to minimize its cross-entropy loss. This minmax optimization is known as \textit{adversarial training} and recently it gained a lot of attention for NLP applications \cite{liu2017adversarial, chen2017adversarial, kim2017adversarial, qin2017adversarial, wu2017adversarial, gui2017part, li2017adversarial, zhang2017adversarial, joty2017cross}. 

%% file: setup.tex
\section{Experimental setup}
\label{sec:setup}

\subsection{Datasets}
\input{tables/data_stats}

For SRL we use the newswire CoNLL-2005 shared task dataset \cite{carreras2005introduction}, annotated with PropBank predicate-argument structures. Sections 2-21 of the WSJ corpus \cite{charniak2000bllip} are used for training and section 24 as dev set. The test set consists of section 23 of WSJ and 3 sections of the Brown corpus.

For ORL we use the manually annotated MPQA 2.0.\ corpus \cite{Wiebe2005AnnotatingEO, wilson2008fine}. It mostly contains news documents, but also travel guides, transcripts of spoken conversations, e-mails, fundraising letters, textbook chapters and translations of Arabic source texts. 

We report detailed pre-processing of MPQA\footnote{Examples how to use our scripts can be found at \url{https://github.com/amarasovic/naacl-mpqa-srl4orl/blob/master/generate_mpqa_jsons.py}.} and data statistics in the Appendix. 

\subsection{Evaluation metrics}\label{setup} 

For both tasks we adopt evaluation metrics from prior work. For SRL, precision is defined as the proportion of semantic roles predicted by a system which are correct, recall is the proportion of gold roles which are predicted by a system, F1 score is the harmonic mean of precision and recall. 

In case of ORL, we report 10-fold CV\footnote{We used the same splits as the prior work \cite{Katiyar2016InvestigatingLF}. We thank the authors for providing the splits.} and repeated 4-fold CV with 
\textit{binary F1 score} and \textit{pro\-po\-rti\-o\-na\-l F1 score}, for holders and targets separately. \textit{Binary precision} is defined as the proportion of predicted holders (targets) that overlap with the gold holder (target), \textit{binary recall} is the proportion of gold holders (targets) for which the model predicts an overlapping holder (target). 
\textit{Proportional recall} measures the proportion of the overlap between a gold holder (target) and an overlapping predicted holder (target), \textit{proportional precision} measures the proportion of the overlap between a predicted holder (target) and an overlapping gold holder (target). F1 scores are the harmonic means of (the corresponding) precision and recall.

\subsection{Training details}
We evaluate our models using two evaluation setting. First, we follow \citet{Katiyar2016InvestigatingLF} which set aside $132$ documents for development and used the remaining 350 documents for 10-fold CV. However, in the 10-fold CV setting, the test sets are more than 3 times smaller than the dev set (Table \ref{tbl:data_stats_1}, row 3), and, consequently, results in high-variance estimates on the test sets. Therefore we additionally evaluate our models with 4-fold CV. We set aside 100 documents for development and use 25\% of the remaining documents for testing. The resulting test sets are comparable in size to the dev set (Table \ref{tbl:data_stats_1}, row 2). We run 4-fold CV twice with two different random seeds.  
We do not tune hyperparameters (HPs), but follow suggestions proposed in the comprehensive HPs study for sequence labeling tasks in \citet{TUD-CS-2017-0196}. All HPs can be found in the Appendix. 

%% file: tables/data_stats.tex
\begin{table}[t]
\centering
\label{data_stats}
\resizebox{\columnwidth}{!}{%
\begin{tabular}{@{}cccccc@{}}
\toprule
 & task & train size & dev size & test size & $|\mathcal{Y}|$  \\
\midrule
CoNLL'05 & SRL & 90750  & 3248 & 6071  & 106  \\
MPQA (4-CV) & ORL & 3141.25 & 1055 & 1036.75 & 7    \\
MPQA (10-CV) & ORL & 3516.3 & 1326 & 349.3 & 7    \\
\bottomrule
\end{tabular}}
\caption{Datasets w/ nb.\ of SRL predicates/ORL opinions in train, dev \& test set, size of label inventory.}
\label{tbl:data_stats_1}
\end{table}

%% file: results.tex
\section{Results}
\label{sec:results}

\input{tables/results_exp1}

\input{tables/results_exp2}

We evaluate all models after every $\ceil*{\frac{\text{train size}}{\text{batch size}}}$ iteration on the ORL dev set and save them if they achieve a higher arithmetic mean of proportional F1 scores of holders and targets on the ORL dev set. The saved models are used for testing. 

We report the mean of F1 scores over $10$ folds and the standard deviation (appears as a subscript) of all models in Table \ref{tbl:results_exp1}. We report the mean of F1 scores over $4$ folds and $2$ different seed (8 evaluations) and the standard deviation of all models in Table \ref{tbl:results_exp2}. Evaluation metrics follow Section \ref{setup}. 

We mark significant difference between MTL models and the single-task (Z\&X-STL) model, observed using a Kolmogorov-Smirnov significance test ($p < 0.05$) \cite{massey1951kolmogorov}, with $\bullet$ in superscript and between the FS-MTL model and other MTL models with $\Diamond$. 

\textbf{STL vs.\ MTL.} In the 10-fold CV evaluation setting (Table \ref{tbl:results_exp1}), the FS-MTL and the H-MTL models improve over the Z\&X-STL model in all evaluation measures, for both holders and targets. When evaluated in the repeated 4-fold CV setting (Table \ref{tbl:results_exp2}), all MTL models improve over the Z\&X-STL model in all evaluation measures, for both holders and targets.

The FS-MTL and the H-MTL models improve \textit{significantly} in all evaluation measures, for both holders and targets, on both dev and test sets, when evaluated with repeated 4-fold CV. With 10-fold CV the improvements are also significant, except for targets on the test set. This is probably due to the small size of the test sets (Table \ref{tbl:data_stats_1}, row 3), which results in a high-variance estimate. Indeed, standard deviations on the 10-fold CV test sets are always much higher compared to the dev set or to the test sets of 4-fold CV. 

It is not surprising that larger improvements are visible in the labeling of holders. They are usually short, less ambiguous and often presented with the A0 semantic role, whereas annotating targets is a challenging task even for humans.\footnote{\citet{wilson2008fine} reports annotator agreement for target labeling of 86.00 binary F1 score.} 

Larger improvements are visible for proportional F1 score than for binary F1 score. That is, more data and SRL knowledge helps the model to better annotate the scope of opinion roles.

\textbf{Comparing MTL models.} In Section \ref{sec:models} we introduced MTL models with task-specific LSTM layers hypothesizing that 
these layers should give MTL models more power to adapt to a variety of potentially problematic cases that we illustrated in the Introduction. However, our results show that the FS-MTL model performs significantly better or comparable to MTL models that include task-specific layers. \citet{TUD-CS-2017-0196} show that MTL is especially sensitive to the selection of HPs. Thus, a firm and solid comparison of the different MTL models requires thorough HP optimization, to properly control the number of parameters and regularization of the models. We leave HP optimization for future work.

%% file: tables/results_exp1.tex
\begin{table*}[t]
\centering
\resizebox{\textwidth}{!}{
\begin{tabular}{@{}c|cc|cc@{}}
\toprule
\multicolumn{1}{c}{} & \multicolumn{4}{c}{\textbf{dev (MPQA)}}       \\
\midrule
\multicolumn{1}{c}{} & \multicolumn{2}{c|}{\textbf{holder}}             & \multicolumn{2}{c}{\textbf{target}}  \\
\midrule
 & \textbf{binary F1}   & \textbf{prop.\ F1}   & \textbf{binary F1}      & \textbf{prop.\ F1}   \\
 \midrule
Z\&X-STL & 80.15$_{1.10}$ &	76.87$_{1.26}$ &	74.62$_{0.67}$	& 70.23$_{1.04}$	 \\
FS-MTL     & 83.68$_{0.44}^{\bullet}$	& 81.45$_{0.58}^{\bullet}$	& \textbf{76.23}$_{0.75}^{\bullet}$	& \textbf{73.01}$_{0.93}^{\bullet}$	 \\
H-MTL      & \textbf{84.14}$_{0.72}^{\bullet}$	& \textbf{81.86}$_{0.48}^{\bullet}$ &	76.11$_{0.61}^{\bullet}$	&72.55$_{0.73}^{\bullet}$	\\

SP-MTL     & 82.18$_{0.89}^{\bullet \Diamond}$	 &79.66$_{0.72}^{\bullet \Diamond}$	 &74.99$_{1.17}^{\Diamond}$	 &71.32$_{1.81}^{\Diamond}$ \\
ASP-MTL    & 82.63$_{0.84}^{\bullet \Diamond}$	& 80.20$_{0.99}^{\bullet \Diamond}$ &	74.24$_{0.58}^{\bullet \Diamond}$ 	& 70.16$_{1.29}^{\Diamond}$ \\
\bottomrule
\end{tabular}
\hspace{0.1cm}
\begin{tabular}{@{}c|cc|cc@{}}
\toprule
\multicolumn{1}{c}{} & \multicolumn{4}{c}{\textbf{test (MPQA)}}        \\
\midrule
\multicolumn{1}{c}{} & \multicolumn{2}{c|}{\textbf{holder}}             & \multicolumn{2}{c}{\textbf{target}}          \\
\midrule
 & \textbf{binary F1}   & \textbf{prop.\ F1}   & \textbf{binary F1}      & \textbf{prop.\ F1}    \\
 \midrule
Z\&X-STL & 80.24$_{2.91}$	& 77.98$_{2.90}$	&76.30$_{2.55}$ &	71.18$_{2.55}$ \\
FS-MTL     & 83.47$_{2.26}^{\bullet}$	& 81.80$_{2.26}^{\bullet}$ &	\textbf{77.60}$_{2.52}$	& \textbf{73.77}$_{2.28}$ \\
H-MTL      & \textbf{84.03}$_{2.65}^{\bullet}$ &	\textbf{82.34}$_{2.51}^{\bullet}$	& 77.41$_{2.14}$	& 73.10$_{1.96}$ \\
SP-MTL     &82.19$_{2.49}^{\bullet}$	 &80.11$_{2.36}^{\bullet \Diamond}$	 &76.01$_{3.03}$	 &71.51$_{3.34}$ \\
ASP-MTL    &	83.15$_{2.92}^{\bullet}$	& 81.12$_{2.66}^{\bullet}$	& 75.89$_{2.66}$ &	71.21$_{2.78}$  \\
\bottomrule
\end{tabular}}
\caption{ORL 10-fold CV results.}
\label{tbl:results_exp1}
\end{table*}

%% file: tables/results_exp2.tex
\begin{table*}[t]
\centering
\resizebox{\textwidth}{!}{
\begin{tabular}{@{}c|cc|cc@{}}
\toprule
\multicolumn{1}{c}{} & \multicolumn{4}{c}{\textbf{dev (MPQA)}}       \\
\midrule
\multicolumn{1}{c}{} & \multicolumn{2}{c|}{\textbf{holder}}             & \multicolumn{2}{c}{\textbf{target}}  \\
\midrule
 & \textbf{binary F1}   & \textbf{prop.\ F1}   & \textbf{binary F1}      & \textbf{prop.\ F1}   \\
 \midrule
Z\&X-STL & 79.73$_{1.19}$ & 77.06$_{1.14}$ & 76.09$_{0.94}$ & 70.45$_{1.07}$  \\
FS-MTL & \textbf{83.58}$_{0.69}^{\bullet}$ & \textbf{82.16}$_{0.59}^{\bullet}$ & \textbf{78.32}$_{1.57}^{\bullet}$ & \textbf{75.09}$_{2.27}^{\bullet}$\\
H-MTL      & 82.36$_{0.81}^{\bullet\Diamond}$ & 80.84$_{0.98}^{\bullet\Diamond}$ & 78.11$_{0.82}^{\bullet}$ & 74.89$_{1.33}^{\bullet}$ \\
SP-MTL     & 82.21$_{0.79}^{\bullet\Diamond}$ & 80.23$_{0.88}^{\bullet\Diamond}$ & 76.14$_{1.18}^{\Diamond}$ & 71.14$_{0.97}^{\Diamond}$ \\
ASP-MTL    & 81.41$_{1.27}^{\Diamond}$ & 79.39$_{1.45}^{\bullet\Diamond}$ & 76.49$_{1.39}$ & 72.13$_{1.87}^{\bullet}$ \\
\bottomrule
\end{tabular}
\hspace{0.1cm}
\begin{tabular}{@{}c|cc|cc@{}}
\toprule
\multicolumn{1}{c}{} & \multicolumn{4}{c}{\textbf{test (MPQA)}}        \\
\midrule
\multicolumn{1}{c}{} & \multicolumn{2}{c|}{\textbf{holder}}             & \multicolumn{2}{c}{\textbf{target}}          \\
\midrule
 & \textbf{binary F1}   & \textbf{prop.\ F1}   & \textbf{binary F1}      & \textbf{prop.\ F1}    \\
 \midrule

Z\&X-STL & 80.42$_{1.92}$ & 77.48$_{2.06}$ & 73.84$_{1.17}$ & 67.03$_{2.13}$ \\
FS-MTL & \textbf{83.67}$_{1.52}^{\bullet}$ & \textbf{81.59}$_{1.50}^{\bullet}$ & 77.04$_{1.45}^{\bullet}$ & 73.01$_{2.53}^{\bullet}$\\
H-MTL      & 82.80$_{1.87}^{\bullet}$ & 80.40$_{1.91}^{\bullet}$ & \textbf{77.12}$_{1.34}^{\bullet}$ & \textbf{73.16}$_{1.78}^{\bullet}$ \\
SP-MTL     &82.51$_{2.17}^{\bullet}$ & 80.03$_{2.00}^{\bullet}$ & 74.61$_{1.32}^{\Diamond}$ & 68.70$_{2.32}^{\Diamond}$ \\
ASP-MTL    &	81.77$_{1.74}^{\Diamond}$ & 79.32$_{1.62}^{\Diamond}$ & 74.92$_{0.84}^{\bullet\Diamond}$ & 69.89$_{1.80}^{\bullet}$\\
\bottomrule
\end{tabular}}
\caption{ORL repeated 4-fold CV results.}
\label{tbl:results_exp2}
\end{table*}

%% file: analysis.tex
\section{Analysis}
\label{sec:analysis}

\input{tables/analysis_true_true_holder}

\input{tables/analysis_true_true_target}

\input{tables/stats_holder}
\input{tables/stats_target}

\input{tables/analysis_false_false_holder}

\input{tables/analysis_false_false_target}

\input{tables/analysis_true_false_holder}

\input{tables/analysis_true_false_target}

Our aim in this section is to analyze what the proposed models are good at, in which ways MTL improves over the single-task ORL model and what could be done to achieve further progress.

We evaluate the FS-MTL and the Z\&X-STL models on the ORL dev set using 4-fold CV repeated twice with different seeds (8 evaluation trials). We say that a model predicts a role of a given opinion expression \textit{correctly} if the model predicts a role that overlaps with the correct role in at least 6 out of 8 evaluation trials. If a model predicts a role that overlaps with the correct role in at most 2 out of 8 trials, we say that the model predicts the role \textit{incorrectly}. The requirement on 6-8 (in)correct predictions reduces the risk of analyzing inconsistent predictions and enables us to draw firmer conclusions. We analyze the following scenarios:

\begin{enumerate}[(i)]
\item both the FS-MTL model and the Z\&X-STL model make correct predictions (Tables \ref{tbl:works_holder}--\ref{tbl:works_target})
\item the FS-MTL model makes a correct prediction, while the Z\&X-STL makes an incorrect prediction (Tables \ref{tbl:improv_holder}--\ref{tbl:improv_target})
\item both models make wrong predictions (Tables \ref{tbl:next_holder}--\ref{tbl:next_target})
\end{enumerate} 

In the following, we categorize predictions in case (i) as \textit{easy cases}, and predictions in case (iii) as \textit{hard cases}. 

In Tables 5-6 and 9--12, the opinion expression is bolded, the correct role is italicized, predictions of the FS-MTL model are colored blue (subscript \textit{FS}), predictions of the Z\&X-STL model are colored yellow (subscript \textit{ZX}) and green marks predictions where both models agree. For simplicity, we show only holders or targets, although the models predict both roles jointly.

\textbf{What works well?} There are 668/1055 instances in the dev set for which both models predict holders correctly, and 663/1055 for targets.

Examples 1--5 in Table \ref{tbl:works_holder} suggest that holders that can be properly labeled by both models (\textit{easy} cases) are subjects of their governing heads or A0 roles. The statistics in Table \ref{tbl:stats_holder} (col.\ 1, rows 2--3) supports this observation.\footnote{The statistics is calculated using the output of mate-tools \cite{bjorkelund2010high}.} In contrast, holders that both models predict incorrectly (\textit{hard} cases) are less frequently subjects or A0 roles (col.\ 2, rows 2--3). Also, \textit{easy} holders are close to the corresponding opinion expression: the average distance is $1.54$ tokens (Table \ref{tbl:stats_holder}, row 4), contrary to the \textit{hard} holders with the average distance of $7.56$.

Examples 1--5 in Table \ref{tbl:works_target} suggest that targets that can be properly labeled by both models are objects of their governing heads or A1 roles. Table \ref{tbl:stats_target}, row 3, shows that the majority of the \textit{easy} targets are indeed A1 roles, in contrast to the \textit{hard} targets. Similar to holders, the \textit{easy} targets are in average $7$ tokens closer to the opinion expression. 

\textbf{What to do for further improvement?} There are 165/1055 instances in the dev set for which both models predict holders incorrectly, and 176 for targets.

As we have seen so far, many holders that are subjects or A0 roles, and targets that are A1 roles, are properly labeled by both models. However, a considerable amount of such holders and targets are not correctly predicted (Table \ref{tbl:stats_holder}--\ref{tbl:stats_target}, col.\ 2, rows 2--3). Thus our models do not work flawlessly for all such cases. A distinguishing property of the \textit{hard} cases is the distance of the role from the opinion. Thus, future work should advance the model' s ability to capture long-range dependencies. 

Examples in Table \ref{tbl:next_holder} demonstrate that holders, harder to label with our models, occur with the corresponding opinions in more complicated syntactic constructions. In the first example, the FS-MTL model does not recognize the possessive and is possibly biased towards picking the country (\textit{Isreal}), which occurs immediately after the opinion. In the second example, the opinion expression is a nominal predicate and the holder is its object. The sentence is in passive voice but the models probably interpret it in the active voice and thus make the wrong prediction. In the third example, the opinion expression is the head of the relative clause that modifies the holder. These examples raise the following questions: would improved
consistency with syntax lead to improvements for ORL and could we train a dependency parsing model with SRL and ORL to help the models handle syntactically harder cases? 

Example 4 shows that holders specific to the MPQA annotation schema are hard to label as they require inference skills:  from \textit{the department said}, we can defeasibly infer that it is \textit{the department} who \textit{expects [this cost] to stand at just \$400/year [...]}. To handle such cases, it would be worth trying training our models jointly with models for recognizing textual entailment. 

Examples 6--7 illustrate that some gap in performance stems from difficulties in processing MPQA. Example 5 has no gold holder, but the models make plausible predictions. For example 6, FS-MTL predicts the discontinuous holder \textit{they ... all}, while MPQA allows only contiguous entities. Therefor our evaluation scripts interpret \textit{they} and \textit{all} as two separate holders and deem \textit{all} as incorrect, resulting with lower precision. Finally, for example 7 our models make plausible predictions. However, the gold holder is always the entity from the coreference cluster that is the closest to the opinion.\footnote{We followed the prior work \cite{Katiyar2016InvestigatingLF}.} The evaluation scripts needs to be extended such that predicting any entity from the coreference cluster is considered to be correct. To conclude, to better evaluate future developments, it would be worth curating MPQA instances with missing roles and extending evaluation 
to account for coreferent holders and discontinuous roles. 

The examples in Table \ref{tbl:next_target} demonstrate that difficulties in labeling targets originate from similar reasons as for holders. Examples 1--3 demonstrate complex syntactic constructions, examples 4--6 MPQA-specific annotations that require inference and example 7 exemplifies a missing target.

\textbf{How does MTL help?} There are 18/1055 instances in the dev set for which the FS model predicts the holder correctly and the Z\&X-STL model does not, and 19/1055 for targets.

For holders, for 9 out 18 of such examples, the Z\&X-STL model does not predict anything (as in Examples 2--5 in Table \ref{tbl:improv_holder}). From Examples 1--5 we notice that SRL data helps to handle more complex syntactic constructions. From Examples 5--7 we observed that using MTL with SRL helps to handle cases when more than one person or organization is present in the close neighborhood of the opinion. For targets, for 11 out of 18 cases the Z\&X-STL model does not predict anything as in Examples 1--2 in Table \ref{tbl:improv_holder}. We conclude that the greatest improvements from the FS-MTL model comes from having far fewer missing roles. 

%% file: tables/analysis_true_true_holder.tex
\begin{table*}[t]
\centering
\resizebox{\textwidth}{!}{%
\resizebox*{!}{\dimexpr\textheight-\lineskip\relax}{%
\centering
\begin{tabular}{lm{19cm}}
\toprule 
1 & \colorbox{bothcolor}{\textit{Malinga}}$_{FS, ZX}$ \textbf{said} according to the guidelines in the booklet, the election had been legitimate . \\
\midrule
2 & movie um-hum that 's interesting so that was a good movie too well do \colorbox{bothcolor}{\textit{you}}$_{FS, ZX}$ \textbf{think} we've covered baseball i think so okay well have a good night \\
\midrule
3 &  \colorbox{bothcolor}{\textit{The nation}}$_{FS, ZX}$  should certainly \textbf{be concerned} about the plans to build a rocket launch pad , work on the infrastructure for which is due to start in 2002 , with launches beginning from 2004 . \\
\midrule
4 & Bam on Sunday said  \colorbox{bothcolor}{\textit{she}}$_{FS, ZX}$ \textbf{believed} Zimbabwe's election was not free and fair , adding they were not in line with international standards as well as those of her organisation . \\
\midrule
5 & The majority report , endorsed only by the ANC , said  \colorbox{bothcolor}{\textit{the observer mission}}$_{FS, ZX}$ \textbf{had noted} that over three million Zimbabweans had cast their votes and this substantially represented the will of the people .\\
\bottomrule
\end{tabular}}
}
\caption{The dev examples for which both models (FS-MTL, Z\&X-STL) \textit{correctly} predict the \textit{holder} in 6/8 trials. 
}
\label{tbl:works_holder}
\end{table*}

%% file: tables/analysis_true_true_target.tex
\begin{table*}[h]
\centering
\resizebox{\textwidth}{!}{%
\resizebox*{!}{\dimexpr\textheight-\lineskip\relax}{%
\centering
\begin{tabular}{lm{19cm}}
\toprule 
1 & Indonesia \textbf{has come under pressure} from \colorbox{fscolor}{several quarters to} \colorbox{bothcolor}{\textit{take tougher action against alleged terrorist leaders} but has} \colorbox{bothcolor}{played down the threat}$_{ZX}$$_{FS}$ .\\
\midrule
2 & Mugabe even talked about his \textbf{desire} to \colorbox{bothcolor}{\textit{keep safeguarding Zimbabwe 's sovereignty and land}}$_{ZX}$ \colorbox{fscolor}{\textit{in spirit}}$_{FS}$ \textit{when he dies} , a dream which the veteran leader said forced him to sacrifice a bright teaching career in the 1950s to lead [...]. \\
\midrule
3 & Under his blueprint , the government \textbf{hopes} to \colorbox{bothcolor}{\textit{stabilize the economy} through curtailing state expenditure , reforming public} \colorbox{bothcolor}{enterprises and expanding agriculture}$_{FS,ZX}$ . \\
\midrule
4 & He said those who thought the election process would be rigged were supporters of the MDC party , adding that they were prejudging and \textbf{wanted} to \colorbox{bothcolor}{\textit{direct the process}}$_{FS,ZX}$ . \\
\midrule 
5 & People in the rural areas support the ruling party because our party has been genuine on \textbf{its policy} on \colorbox{bothcolor}{\textit{land reform}}$_{FS,ZX}$.\\
\bottomrule
\end{tabular}}
}
\caption{The dev examples for which both models (FS-MTL, Z\&X-STL) \textit{correctly} predict the \textit{target} in 6/8 trials. 
}
\label{tbl:works_target}
\end{table*}

%% file: tables/stats_holder.tex
\begin{table}[t]
\centering
\resizebox{0.48\textwidth}{!}{%
\begin{tabular}{lcc}
\toprule 
					   & easy & hard\\
\midrule                      
\% opinions that are predicates  & 91.32 & 93.33 \\
\% holders that are subjects   & 77.84 & 38.79 \\
\% holders that are A0 roles         & 74.10 & 33.33 \\
avg.\ distance between holders \& opinions & 1.54  & 7.56 \\
\bottomrule
\end{tabular}}
\caption{Statistics of \textit{holder} prediction.}
\label{tbl:stats_holder}
\end{table}

%% file: tables/stats_target.tex
\begin{table}[t]
\centering
\resizebox{0.48\textwidth}{!}{%
\begin{tabular}{lcc}
\toprule 
					   & easy & hard\\
\midrule                      
\% opinions that are predicates  & 92.58 &	89.20 \\
\% target's heads that are objects   & 22.12 &	14.77 \\
\% targets that are A1 roles         & 70.62	& 42.61 \\
\% targets that are A2 roles         & 9.00 &	0.57 \\
avg.\ distance between targets \& opinions &2.29 &	8.46 \\
\bottomrule
\end{tabular}}
\caption{Statistics of \textit{target} prediction.}
\label{tbl:stats_target}
\end{table}

%% file: tables/analysis_false_false_holder.tex
\begin{table*}[t]
\centering
\resizebox{\textwidth}{!}{%
\resizebox*{!}{\dimexpr\textheight-\lineskip\relax}{%
\centering
\begin{tabular}{lm{19cm}}
\toprule 
1 & It would be entirely improper if , in \textit{its} \textbf{defense of} \colorbox{fscolor}{Israel}$_{FS}$ , the United States continues to exert pressure on [...] .  \\
\midrule
2 & \colorbox{bothcolor}{Indonesia}$_{FS, ZX}$ \textbf{has come under pressure} from \textit{several quarters} to take tougher action against alleged terrorist leaders but has played down the threat . \\
\midrule
3 & Australia should adhere to the \textit{Cardinal Principle of International Law} , which \textbf{states} that all nations in the world must first respect and promote the humanitarian interests and progress of all humankind . \\
\midrule
4 & \textit{The department} said that it will cost \$ 600 for an HIV/AIDS patient per year at this time , and the following years this cost is \textbf{expected} to stand at just \$ 400/year for one patient as the production of such drugs becomes stable . \\
\midrule
5 & \colorbox{stlcolor}{The Organisation of African Unity OAU}$_{ZX}$  also backed Zimbabwean President Robert Mugabe 's re-election , with  \colorbox{bothcolor}{its observer team}$_{FS, ZX}$  \textbf{describing} the poll as " transparent , credible , free and fair " .  \\
\midrule
6 & Regarding the American proposed Anti-Missile Defense System too , neither Russia , China , Japan , nor even the European Union , had shown any enthusiasm ; rather \colorbox{fscolor}{\textit{they}}$_{FS}$  had \colorbox{bothcolor}{all}$_{FS,ZX}$  \textbf{expressed} their reserves on the project . \\
\midrule
7 & The president renewed his pledge to thwart \colorbox{bothcolor}{terrorist groups}$_{FS,ZX}$ \textit{who} \textbf{want} to " mate up " with regimes hoping to acquire weapons of mass destruction and said " nations will come with us " if the US-led war on terrorism is extended . \\
\bottomrule
\end{tabular}}
}
\caption{The dev examples for which both models (FS-MTL, Z\&X-STL) \textit{incorrectly} predict the \textit{holder} in 6/8 trials.}
\label{tbl:next_holder}
\end{table*}

%% file: tables/analysis_false_false_target.tex
\begin{table*}[!h]
\centering
\resizebox{\textwidth}{!}{%
\resizebox*{!}{\dimexpr\textheight-\lineskip\relax}{%
\centering
\begin{tabular}{lm{19cm}}
\toprule 
1 & \textit{State-sanctioned land invasions} , several times \textbf{declared} illegal by Zimbabwe 's courts , as well as a drought have disrupted Zimbabwe 's food production and famine is already looming in much of the country . \\
\midrule
2 & But he \textbf{told} \colorbox{bothcolor}{the nation}$_{FS,ZX}$ that in spite of \textit{stiff opposition to the agrarian reforms from powerful Western countries ,} \textit{especially the country 's former colonial power of Britain} , he would press ahead to seize farms from whites and [...] . \\
\midrule
3 & If the Europeans wish to influence Israel in the political arena -- in \textit{a direction} that many in Israel \textbf{would support wholeheartedly} -- they will not be able to promote their positions in such a manner .\\
\midrule
4 & \colorbox{bothcolor}{They}$_{FS,ZX}$ are fully aware that these are dangerous \textit{individuals} , he \textbf{said} during a press conference [...] . \\
\midrule
5 & And her little girl \textbf{just complained} , " I don't want to \textit{wash the dishes} " .\\
\midrule 
6 & During \textit{President Bush's speech} , I \textbf{thought} of \colorbox{stlcolor}{heckling}$_{ZX}$ ; '\colorbox{fscolor}{What are you going to do with the Kyoto Protocol ?}$_{FS}$' \\
\midrule 
7 & At first I \textbf{didn't want} \colorbox{stlcolor}{to} \colorbox{bothcolor}{apply for it}$_{FS,ZX}$ , but the principal called me during the summer months and said , " Sandra the time is running out , you need to apply ". \\
\bottomrule
\end{tabular}}
}
\caption{The dev examples for which both models (FS-MTL, Z\&X-STL) \textit{incorrectly} predict the \textit{target} in 6/8 trials.}
\label{tbl:next_target}
\end{table*}

%% file: tables/analysis_true_false_holder.tex
\begin{table*}[t]
\centering
\resizebox{\textwidth}{!}{%
\resizebox*{!}{\dimexpr\textheight-\lineskip\relax}{%
\centering
\begin{tabular}{lm{19cm}}
\toprule 
1 & Yoshihisa Murasawa , a management consultant for Booz-Allen \& Hamilton Japan Inc. , said \colorbox{bothcolor}{\textit{his firm}}$_{FS,ZX}$ will likely be recommending acquisitions of \colorbox{stlcolor}{Japanese companies more}$_{ZX}$ often to foreign clients in the future .\\
\midrule
2 & \colorbox{fscolor}{\textit{The source}}$_{FS}$ , interviewed by Interfax in Grozny , \textbf{expressed confidence} that that the command of the Russian forces in Chechnya would soon `` be able to obtain documentary confirmation '' that Khattab was dead . \\
\midrule
3 & \colorbox{fscolor}{\textit{The Commonwealth team earlier} this week}$_{FS}$ \textbf{said} that " the conditions in Zimbabwe did not adequately allow the free and fair expression of will by the electorate ". \\
\midrule
4 & Publishing such biased reports will only create \textbf{mistrust} among \colorbox{fscolor}{nations}$_{FS}$ regarding the objectives and independence of the UN Commission on Human Rights . \\
\midrule
5 & The Inkatha Freedom Party , \colorbox{stlcolor}{Democratic Alliance , New National Party , African Christian Democratic Party , the Pan} \colorbox{stlcolor}{Africanist Congress and the United Christian Democratic Party}$_{ZX}$ had disagreed with \colorbox{fscolor}{the ANC}$_{FS}$ \textbf{conclusion} . \\
\midrule
6 & The Nigerian leader ,  \colorbox{stlcolor}{President Olusegun Obasanjo}$_{ZX}$  , had urged the  \colorbox{bothcolor}{\textit{minister}}$_{FS,ZX}$  not to attack Blair frontally over Britain 's negative position regarding Zimbabwe , but to deal [...] . \\
\midrule
7 & \colorbox{stlcolor}{US diplomats}$_{ZX}$ say \colorbox{bothcolor}{\textit{Bush}}$_{FS,ZX}$ \textbf{will seek to support} Kim 's Nobel Prize winning policy by offering new talks with the North , while remaining firm about North Korea 's missile sales and its feared chemical and biological weapons programmes. \\
\bottomrule
\end{tabular}}
}
\caption{The dev examples for which the FS-MTL model \textit{correctly} predicts the \textit{holder} in 6/8 trials, whereas the Z\&X-STL model predicts \textit{incorrectly} in 6/8 trials. 
}
\label{tbl:improv_holder}
\end{table*}

%% file: tables/analysis_true_false_target.tex
\begin{table*}[t]
\centering
\resizebox{\textwidth}{!}{%
\resizebox*{!}{\dimexpr\textheight-\lineskip\relax}{%
\centering
\begin{tabular}{lm{19cm}}
\toprule 
1 & In most cases he \textbf{described} \colorbox{fscolor}{\textit{the legal punishments}}$_{FS}$ \textit{like floggings and executions of murderers and major drug traffickers} \textit{that are applied based on the Shria , or Islamic law as human rights violations }. \\
\midrule
2 & In another \textbf{verbal attack} Kharazi accused \colorbox{fscolor}{\textit{the United States}}$_{FS}$ of wanting to exercise " world dictatorship " since the " horrible attacks " of September 11 . \\
\midrule
3 & He said those who thought the election process would be rigged were supporters of the MDC party , adding that they \textbf{were prejudging} and \colorbox{stlcolor}{wanted to direct the process}$_{ZX}$ . \\
\midrule 
4 & However , the fact that certain countries \colorbox{stlcolor}{have a more balanced view of the conflict}$_{ZX}$ is not the only reason to doubt that \colorbox{fscolor}{\textit{anti-Israeli decisions}}$_{FS}$ will , in fact , \textbf{be adopted} . \\
\midrule
5 & But \colorbox{fscolor}{\textit{his tough stand on P'yongyang}}$_{FS}$ \textbf{has provoked concern} in \colorbox{stlcolor}{Seoul}$_{ZX}$ , where President Kim Tae-chung , who is in the last year of his five-year term , has been trying to prise the hermit state out of isolation . \\
\bottomrule
\end{tabular}}
}
\caption{The dev examples for which the FS-MTL model \textit{correctly} predicts the \textit{target} in 6/8 trials, whereas the Z\&X-STL model predicts \textit{incorrectly} in 6/8 trials. 
}
\label{tbl:improv_target}
\end{table*}

%% file: related.tex
\section{Related work}
\label{sec:related}

\textbf{FGOA.} Closest to our work are \citet{Yang2013JointIF} (Y\&C) and \citet{Katiyar2016InvestigatingLF} (K\&C). They as well label both holders and targets in MPQA. By contrast, our focus is on the task of ORL. We thus refrain from predicting opinion expressions first, to ensure a reproducible evaluation setup on a fixed set of gold opinion expressions. The MTL models we develop in this work will, however, be the basis for the full task in a later stage. Because of these differences, direct comparison to Y\&C and K\&C is not possible. However, 
if we compare our results we notice a big gap that demonstrates that opinion expression extraction is the import step in FGOA. Similar to K\&C, \citet{liu-joty-meng:2015:EMNLP} jointly labels opinion expressions and their targets in reviews.

Some work focuses entirely on labeling of opinion expressions \cite{yang2014joint, irsoy-cardie:2014:EMNLP2014}. Other work looks into specific subcategories of ORL: opinion role induction for verbal predicates \cite{wiegand-ruppenhofer:2015:CoNLL}, categorization of opinion words into actor and speaker view \cite{wiegand-schulder-ruppenhofer:2016:N16-1}, opinion roles extraction on opinion compounds \cite{wiegand-bocionek-ruppenhofer:2016:N16-1}. \citet{wiegand-ruppenhofer:2015:CoNLL} report $72.54$ binary F1 score for labeling of holders in MPQA (results for targets are not reported). 

\textbf{Neural SRL.} New neural SRL models have emerged \cite{he-EtAl:2017:Long3, yang-mitchell:2017:EMNLP2017, marcheggiani-titov:2017:EMNLP2017} since we started this work. In future work we can improve our models with such new proposals. 

\textbf{Auxiliary tasks for MTL.} Other work investigates under which conditions MTL is effective. \citet{martinezalonso-plank:2017:EACLlong} show that the best auxiliary tasks have low kurtosis of labels (usually a small label set) and high entropy (labels occur uniformly). We show that the best MTL model for ORL is the model which uses shared layers only. Thus it seems reasonable to consider only a small and uniform SRL label set \{A0, A1\}. 

\citet{bingel2017identifying} show that MTL works when the main task has a flattening learning curve, but the auxiliary task curve is still steep. We notice such  behavior in our learning curves.

%% file: conclusions.tex
\section{Conclusions}

We address the problem of scarcity of annotated training data for labeling of opinion holders and targets (ORL) using multi-task learning (MTL) with Semantic Role Labeling (SRL). We adapted a recently proposed neural SRL model for ORL and enhanced it with different MTL techniques. Two MTL models achieve significant improvements with all evaluation measures, for both holders and targets, on both dev and test set, when evaluated with repeated 4-fold CV. We recommend evaluation with comparable dev and test set sizes for future work, as this enables more reliable evaluation. 

With deeper analysis we show that future developments should improve the ability of the models to capture long-range dependencies, investigate if consistency with syntax can improve ORL, and consider other auxiliary tasks such as dependency parsing or recognizing textual entailment. We emphasize that future improvements can be measured more reliably if the evaluation covers opinion expressions with missing roles and considers all mentions in opinion role coreference chains as well as discontinuous roles.

%% file: supplement.tex
\section{MPQA Pre-processing}

\input{tables/stats_4fold.tex}

\input{tables/stats_10fold.tex}

MPQA is challenging not only because it captures a variety of phenomena as we have illustrated in the Introduction, but as well because it is hard to process it in such a way that it can be presented to a neural sequence labeling model. Code or sufficient description how the corpus was processed is not available from the prior work. 

The first difficulty is that we are designing a model that labels at the token-level, but annotation spans are given in bytes. Thus, we used Stanford CoreNLP \cite{manning-EtAl:2014:P14-5} which tokenizes text and gives the byte span of every token.\footnote{We used python wrapper: \url{https://github.com/brendano/stanford_corenlp_pywrapper}} However, due to to the absence of punctuation for transcripts of spoken conversations the sentence splitter  treats a whole document as if it were one sentence. Therefore, for sentences longer than $150$ tokens, we take $15$ tokens preceding the opinion expression, the expression itself and $15$ tokens after as proxy for a sentence that we present to the model.

Opinion expressions we are interested in are annotated in MPQA as \textit{direct subjectives} (DSEs). We discard implicit DSEs  which frequently point to the attitude which covers the whole sentence and reflects the attitude of the author of the document as in example (6).  These DSEs are not useful for the task we are looking into.
Although such DSEs should be marked with the \texttt{implicit} attribute, sometimes they are not. Some of such cases we capture by demanding that a DSE is longer than one byte and that the author is not the only holder. There are few DSEs for which byte spans did not match with any sentence, and we discard those as well. 

\enumsentence{But there can not be any real [\textbf{talk}]$_O$ of success until the broad strategy against terrorism begins to bear fruit.}

For every document, we collected from the corresponding annotation file:
identifiers and byte spans of all holders marked with \texttt{GATE\_agent} ($\mathcal{H}$), attitudes marked with \texttt{GATE\_attitude}, and targets marked with \texttt{GATE\_target}. Holders and targets can be marked multiple times with the same id, but with different byte spans. If the \texttt{nested-source} attribute of a DSE or the \texttt{target-link} attribute of its attitude point to identifiers of such holders and targets, we pick the byte spans which are closest to the DSE. In many cases the \texttt{nested-source} attribute of a DSE pointed to a holder which is not marked in the annotation file ($\notin \mathcal{H}$). We tried to fix the \texttt{nested-source} attribute by doing the following transformations: (1) adding 'w' to the beginning (e.g.\ nhs $\mapsto$ w, nhs), (2) removing 'w' from the beginning (e.g.\ w, ip $\mapsto$ ip), (3) removing duplicates (e.g.\ w, mug, mug $\mapsto$ w, mug). Although these transformations helped a lot, they are a few holders and targets we could not trace.

In some cases, as in example (7), an opinion expression and its opinion roles overlap. In average, we discard 74.7 such holders and 16.2 targets, because we train the output CRF to predict only one label by token. Notice that the prior work \cite{Katiyar2016InvestigatingLF} had to do the same.

\enumsentence{Mugabe said [Zimbabwe]$_T$ needed their continued support against what he called [\textbf{hostile [international]$_H$ attention}]$_O$.}

We discard inferred attitudes, as labeling of their targets is considered to be another task \cite{deng-choi-wiebe:2013:Short, deng-wiebe:2014:EACL, RUPPENHOFER16.832}.

Further, a DSE can have multiple attitudes and each attitude can point to different targets. Again, because the model can predict only one label by
token, we have to pick one attitude and non-overlapping targets. We chose attitudes according to the following priorities: sentiment, intention, agreement, arguing, other-attitude, speculation. 

We kept DSEs with the \texttt{insubstantial} attribute which are either not significant (8) or not not real within the discourse (9). Our models should demonstrate the ability of properly labeling roles of insubstantial DSEs. However, note that when FGOA is used for opinion-oriented summarization or QA, opinion roles of insubstantial opinions should not be labeled. A full FGOA system should additionally predict whether an opinion is substantial within the discourse, before labeling its opinion roles.

\enumsentence{[...] it completely supports the [U.S.]$_H$ [\textbf{stance}]$_O$ [...].} 

\enumsentence{[...] Antonio Martino, meanwhile, said [...] that his country would not support an attack on Iraq without "proven proof" that [Baghdad]$_H$ is [\textbf{supporting}]$_O$ [al Qaeda]$_T$.} 

Finally, DSE, holder and target annotations allow an attribute that indicates whether an annotator was uncertain with possible values: somewhat- and very-uncertain. We did not discard those believing that they would have been discarded by the corpus creators if they are really incorrect.

For reproducibility we report detailed data statistics in Tables \ref{tbl:stats_4fold} and \ref{tbl:stats_10fold}: average number (calculated over folds) of all extracted DSEs, implicit DSEs, inferred DSEs, DSEs used in experiments (not implicit or inferred), somewhat uncertain DSEs used in experiments, very uncertain DSEs used in experiments, insubstantial DSEs used in experiments, the average number (calculated over folds) of DSEs used in experiments without a holder, without a target, without the \texttt{attitude-link} attribute, without both roles, the average number (calculated over folds) of holders, somewhat uncertain holders, very uncertain holders, targets, somewhat uncertain targets and very uncertain targets, the average number (calculated over folds) of different attitude types used in the experiments.

Examples how to easily use our MPQA pre-processing scripts can be found at \url{https://github.com/amarasovic/naacl-mpqa-srl4orl/blob/master/mpqa2-pytools.ipynb}. 

\section{Training details}

The code for training and evaluating our models can be found at \url{https://github.com/amarasovic/naacl-mpqa-srl4orl}. 

\textbf{Input representation.} We used 100d GloVe word embeddings \cite{pennington2014glove} pre-trained on Gigaword and Wikipedia and did not fine-tune them. For MTL models vocabulary was built from all the words in the training data of both tasks, and OOV words were replaced with an UNK token. The embedding of the context of a predicate or an opinion is the average of the embeddings of the predicate or the opinion phrase, of 2 preceding words and 2 words after. 

\textbf{Weights initialization.} The size of all LSTM hidden states was set to $100$. The number of the backward and the forward LSTM layers is set to $3$, which counts for 6 LSTM layers in Z\&X. Z\&X achieved circa  2\% higher SRL F1 score with 8 LSTM layers, but such a deep model would cause overfitting on the small-sized ORL data. In the H-MTL model, SRL is supervised at the 2nd LSTM layer. We initialized the LSTM weights with random orthogonal matrices \cite{henaff2016orthogonal}, all other weight matrices with the \textit{He initialization} \cite{he2015delving}. LSTM forget biases were initialized with 1s \cite{jozefowicz2015empirical}, all other biases with 0s. 

\textbf{Optimization.} We trained our model in mini-batches of size $32$ using Adam \cite{kingma2014adam} with the learning rate of $10^{-3}$. For MTL we alternate batches from different tasks. We clip gradients by global norm \cite{pascanu2013difficulty}, with a clipping value set to $1$. Single-task models were trained for 10K iterations and MTL models for 20K. One epoch counts for $\ceil*{\frac{\text{train size}}{\text{batch size}}}$ iterations. We stop training if the arithmetic mean of proportional F1 scores of holders and targets is not improved in $25$ epochs. For the minmax optimization we use a gradient reversal layer \cite{ganin2015unsupervised}. The discriminator's cross-entropy loss is scaled with $0.1$.

\textbf{Regularization.} Variational dropout \cite{Gal2016ATG} with a keep probability $k_p \in 0.85$ was applied to the outputs and the recurrent connections of the LSTMs. Standard dropout \cite{srivastava2014dropout} was applied to the output classifier weights with a keep probability $k_p \in 0.85$ and to the input embeddings with $k_p \in 0.7$.

%% file: tables/stats_4fold.tex
\begin{table*}[t]
\centering
\resizebox{\textwidth}{!}{%
\resizebox*{!}{\dimexpr\textheight-\lineskip\relax}{%
\centering
\begin{tabular}{cccccc}
\toprule 
          & \# \textbf{DSEs (incl.\ ignored)}  & \textbf{\# implicit DSEs (ignored)} & \textbf{\# inferred DSEs (ignored)} & \textbf{\# filtered DSEs}  & \textbf{\# some uncrt. filt.\ DSEs}   \\
\midrule 
TRAIN (avg) & 3723.5                       & 481                        & 101.25                     & 3141.25                    & 133                            \\
TEST (avg)  & 1229.5                       & 159                        & 33.75                      & 1036.75                    & 44                             \\
DEV       & 1263                         & 168                        & 40                         & 1055                       & 43                             \\
\midrule
           & \textbf{\# very uncrt. filt.\ DSEs} & \textbf{\# no Hs filt.\ DSEs} & \textbf{\# no roles filt.\ DSEs}  & \textbf{\# no Ts filt.\ DSEs} & \textbf{\# insubs.\ filt.\ DSEs} \\
\midrule   
TRAIN (avg) & 40.5                         & 171.25                     & 66.75                      & 413.75                     & 528.75                         \\
 TEST (avg)  & 13.5                         & 56.75                      & 22.25                      & 136.25                     & 174.25                         \\
 DEV       & 15                           & 56                         & 22                         & 146                        & 180                            \\
 \midrule
 		  & \textbf{\# Hs of filt.\ DSEs}          & \textbf{\# Ts  of filt.\ DSEs}        & \textbf{\# some uncrt. Hs}          & \# \textbf{some uncrt. Ts}          & \textbf{\# overlap. entites}            \\
\midrule   
TRAIN (avg) & 2903.25                      & 19528.5                    & 17.25                      & 27.75                      & 961                            \\
TEST (avg)  & 957.75                       & 6424.5                     & 5.75                       & 9.25                       & 318                            \\
DEV       & 977                          & 6073                       & 5                          & 6                          & 305                            \\
\midrule
          & \textbf{sentiment neg}                & \textbf{sentiment pos}              & \textbf{arguing pos}                & \textbf{other attitude}             & \textbf{intention pos}                  \\
\midrule  
TRAIN (avg) & 946                          & 817.5                      & 438.25                     & 381                        & 238.75                         \\
 TEST (avg)  & 314                          & 270.5                      & 143.75                     & 126                        & 79.25                          \\
 DEV       & 299                          & 300                        & 131                        & 126                        & 66                             \\
\midrule
           & \textbf{arguing neg}                  & \textbf{agree pos}                  & \textbf{speculation}                & \textbf{agree neg}                  & \textbf{intention neg}                 \\
 \midrule   
TRAIN (avg) & 110.5                        & 99.25                      & 64.5                       & 68.25                      & 19.5                           \\
TEST (avg)   & 35.5                         & 32.75                      & 20.5                       & 22.75                      & 6.5                            \\
DEV        & 48                           & 40                         & 25                         & 31                         & 5    \\
\bottomrule
\end{tabular}}}
\caption{Statistics of the ORL (MPQA) data for 4-fold CV.}
\label{tbl:stats_4fold}
\end{table*}

%% file: tables/stats_10fold.tex
\begin{table*}[t]
\centering
\resizebox{\textwidth}{!}{%
\resizebox*{!}{\dimexpr\textheight-\lineskip\relax}{%
\centering
\begin{tabular}{cccccc}
\toprule 
          & \# \textbf{DSEs (incl.\ ignored)}  & \textbf{\# implicit DSEs (ignored)} & \textbf{\# inferred DSEs (ignored)} & \textbf{\# filtered DSEs}  & \textbf{\# some uncrt. filt.\ DSEs}   \\
\midrule          
TRAIN (avg)                 & 4173.3                       & 537.3                      & 119.7                     & 3516.3                       & 137.7                          \\
TEST (avg)                  & 457.8                        & 43.9                       & 29.9                      & 349.3                        & 15.2                           \\
DEV                         & 1579                         & 211                        & 42                        & 1326                         & 67   \\                          \\
\midrule
           & \textbf{\# very uncrt. filt.\ DSEs} & \textbf{\# no Hs filt.\ DSEs} & \textbf{\# no roles filt.\ DSEs}  & \textbf{\# no Ts filt.\ DSEs} & \textbf{\# insubs.\ filt.\ DSEs} \\
\midrule               
TRAIN (avg)                 & 47.7                         & 187.2                      & 77.4                      & 459.9                        & 567.9                          \\
TEST (avg)                  & 7.3                          & 19.3                       & 11.8                      & 82.6                         & 150.5                          \\
DEV                         & 16                           & 76                         & 25                        & 185                          & 252   \\                         
 \midrule
 		  & \textbf{\# Hs of filt.\ DSEs}          & \textbf{\# Ts  of filt.\ DSEs}        & \textbf{\# some uncrt. Hs}          & \# \textbf{some uncrt. Ts}          & \textbf{\# overlap. entites}            \\
\midrule              
TRAIN (avg)                 & 3251.7                       & 21664.8                    & 17.1                      & 27.9                         & 1064.7                         \\
TEST (avg)                  & 957.4                        & 1700                       & 19.4                      & 37.8                         & 84.9                           \\
DEV                         & 1225                         & 7978                       & 9                         & 12                           & 401 \\
\midrule
           & \textbf{sentiment neg}                & \textbf{sentiment pos}              & \textbf{arguing pos}                & \textbf{other attitude}             & \textbf{intention pos}                  \\
\midrule               
TRAIN (avg)                 & 1008.9                       & 949.5                      & 471.6                     & 440.1                        & 266.4                          \\
TEST (avg)                  & 107.8                        & 89.4                       & 50.7                      & 40.2                         & 25.6                           \\
DEV                         & 438                          & 333                        & 189                       & 144                          & 88 \\
\midrule
           & \textbf{arguing neg}                  & \textbf{agree pos}                  & \textbf{speculation}                & \textbf{agree neg}                  & \textbf{intention neg}                 \\
 \midrule              
TRAIN (avg)                 & 133.2                        & 115.2                      & 80.1                      & 74.7                         & 16.2                           \\
TEST (avg)                  & 14.1                         & 11.1                       & 8.6                       & 6.5                          & 1.428571429                    \\
DEV                         & 46                           & 44                         & 21                        & 39                           & 13  \\
\bottomrule
\end{tabular}}}
\caption{Statistics of the ORL (MPQA) data for 10-fold CV.}
\label{tbl:stats_10fold}
\end{table*}

%% file: camera-ready-srl4orl.bbl
\begin{thebibliography}{}
\expandafter\ifx\csname natexlab\endcsname\relax\def\natexlab#1{#1}\fi

\bibitem[{Bingel and S{\o}gaard(2017)}]{bingel2017identifying}
Joachim Bingel and Anders S{\o}gaard. 2017.
\newblock \href{http://www.aclweb.org/anthology/E17-2026}{Identifying
  beneficial task relations for multi-task learning in deep neural networks}.
\newblock In {\em Proceedings of the 15th Conference of the European Chapter of
  the Association for Computational Linguistics: Volume 2, Short Papers\/}.
  Association for Computational Linguistics, Valencia, Spain, pages 164--169.
\newblock \url{http://www.aclweb.org/anthology/E17-2026}.

\bibitem[{Bj{\"o}rkelund et~al.(2010)Bj{\"o}rkelund, Bohnet, Hafdell, and
  Nugues}]{bjorkelund2010high}
Anders Bj{\"o}rkelund, Bernd Bohnet, Love Hafdell, and Pierre Nugues. 2010.
\newblock A high-performance syntactic and semantic dependency parser.
\newblock In {\em Proceedings of the 23rd International Conference on
  Computational Linguistics: Demonstrations\/}. Association for Computational
  Linguistics, pages 33--36.

\bibitem[{Carreras and M{\`a}rquez(2005)}]{carreras2005introduction}
Xavier Carreras and Llu{\'i}s M{\`a}rquez. 2005.
\newblock \href{http://www.aclweb.org/anthology/W/W05/W05-0620}{{Introduction
  to the {CoNLL}-2005 Shared Task: Semantic Role Labeling}}.
\newblock In {\em Proceedings of the Ninth Conference on Computational Natural
  Language Learning (CoNLL-2005)\/}. Association for Computational Linguistics,
  Ann Arbor, Michigan, pages 152--164.
\newblock \url{http://www.aclweb.org/anthology/W/W05/W05-0620}.

\bibitem[{Charniak et~al.(2000)Charniak, Blaheta, Ge, Hall, Hale, and
  Johnson}]{charniak2000bllip}
Eugene Charniak, Don Blaheta, Niyu Ge, Keith Hall, John Hale, and Mark Johnson.
  2000.
\newblock Bllip 1987-89 wsj corpus release 1.
\newblock {\em Linguistic Data Consortium, Philadelphia\/} 36.

\bibitem[{Chen et~al.(2017)Chen, Shi, Qiu, and Huang}]{chen2017adversarial}
Xinchi Chen, Zhan Shi, Xipeng Qiu, and Xuanjing Huang. 2017.
\newblock \href{http://aclweb.org/anthology/P17-1110}{Adversarial
  multi-criteria learning for chinese word segmentation}.
\newblock In {\em Proceedings of the 55th Annual Meeting of the Association for
  Computational Linguistics (Volume 1: Long Papers)\/}. Association for
  Computational Linguistics, Vancouver, Canada, pages 1193--1203.
\newblock \url{http://aclweb.org/anthology/P17-1110}.

\bibitem[{Choi et~al.(2006)Choi, Breck, and Cardie}]{choi2006joint}
Yejin Choi, Eric Breck, and Claire Cardie. 2006.
\newblock \href{http://www.aclweb.org/anthology/W/W06/W06-1651}{Joint
  extraction of entities and relations for opinion recognition}.
\newblock In {\em Proceedings of the 2006 Conference on Empirical Methods in
  Natural Language Processing\/}. Association for Computational Linguistics,
  Sydney, Australia, pages 431--439.
\newblock \url{http://www.aclweb.org/anthology/W/W06/W06-1651}.

\bibitem[{Deng et~al.(2013)Deng, Choi, and Wiebe}]{deng-choi-wiebe:2013:Short}
Lingjia Deng, Yoonjung Choi, and Janyce Wiebe. 2013.
\newblock
  \href{http://www.aclweb.org/anthology/P13-2022}{Benefactive/malefactive event
  and writer attitude annotation}.
\newblock In {\em Proceedings of the 51st Annual Meeting of the Association for
  Computational Linguistics (Volume 2: Short Papers)\/}. Association for
  Computational Linguistics, Sofia, Bulgaria, pages 120--125.
\newblock \url{http://www.aclweb.org/anthology/P13-2022}.

\bibitem[{Deng and Wiebe(2014)}]{deng-wiebe:2014:EACL}
Lingjia Deng and Janyce Wiebe. 2014.
\newblock \href{http://www.aclweb.org/anthology/E14-1040}{Sentiment propagation
  via implicature constraints}.
\newblock In {\em Proceedings of the 14th Conference of the European Chapter of
  the Association for Computational Linguistics\/}. Association for
  Computational Linguistics, Gothenburg, Sweden, pages 377--385.
\newblock \url{http://www.aclweb.org/anthology/E14-1040}.

\bibitem[{Gal and Ghahramani(2016)}]{Gal2016ATG}
Yarin Gal and Zoubin J.~C. Ghahramani. 2016.
\newblock A theoretically grounded application of dropout in recurrent neural
  networks.
\newblock In {\em Advances in Neural Information Processing Systems 29: Annual
  Conference on Neural Information Processing Systems (NIPS)\/}.

\bibitem[{Ganin and Lempitsky(2015)}]{ganin2015unsupervised}
Yaroslav Ganin and Victor Lempitsky. 2015.
\newblock Unsupervised domain adaptation by backpropagation.
\newblock In {\em International Conference on Machine Learning (ICML)\/}. pages
  1180--1189.

\bibitem[{Gui et~al.(2017)Gui, Zhang, Huang, Peng, and Huang}]{gui2017part}
Tao Gui, Qi~Zhang, Haoran Huang, Minlong Peng, and Xuanjing Huang. 2017.
\newblock \href{https://www.aclweb.org/anthology/D17-1255}{Part-of-speech
  tagging for twitter with adversarial neural networks}.
\newblock In {\em Proceedings of the 2017 Conference on Empirical Methods in
  Natural Language Processing\/}. Association for Computational Linguistics,
  Copenhagen, Denmark, pages 2401--2410.
\newblock \url{https://www.aclweb.org/anthology/D17-1255}.

\bibitem[{Hashimoto et~al.(2017)Hashimoto, Xiong, Tsuruoka, and
  Socher}]{hashimoto2016joint}
Kazuma Hashimoto, Caiming Xiong, Yoshimasa Tsuruoka, and Richard Socher. 2017.
\newblock \href{https://www.aclweb.org/anthology/D17-1046}{A joint many-task
  model: Growing a neural network for multiple nlp tasks}.
\newblock In {\em Proceedings of the 2017 Conference on Empirical Methods in
  Natural Language Processing\/}. Association for Computational Linguistics,
  Copenhagen, Denmark, pages 446--456.
\newblock \url{https://www.aclweb.org/anthology/D17-1046}.

\bibitem[{He et~al.(2015)He, Zhang, Ren, and Sun}]{he2015delving}
Kaiming He, Xiangyu Zhang, Shaoqing Ren, and Jian Sun. 2015.
\newblock {Delving deep into rectifiers: Surpassing human-level performance on
  imagenet classification}.
\newblock In {\em Proceedings of the IEEE international conference on computer
  vision\/}. pages 1026--1034.

\bibitem[{He et~al.(2017)He, Lee, Lewis, and Zettlemoyer}]{he-EtAl:2017:Long3}
Luheng He, Kenton Lee, Mike Lewis, and Luke Zettlemoyer. 2017.
\newblock \href{http://aclweb.org/anthology/P17-1044}{Deep semantic role
  labeling: What works and what's next}.
\newblock In {\em Proceedings of the 55th Annual Meeting of the Association for
  Computational Linguistics (Volume 1: Long Papers)\/}. Association for
  Computational Linguistics, Vancouver, Canada, pages 473--483.
\newblock \url{http://aclweb.org/anthology/P17-1044}.

\bibitem[{Henaff et~al.(2016)Henaff, Szlam, and LeCun}]{henaff2016orthogonal}
Mikael Henaff, Arthur Szlam, and Yann LeCun. 2016.
\newblock {Recurrent orthogonal networks and long-memory tasks}.
\newblock In {\em Proceedings of the 33rd International Conference on Machine
  Learning (ICML)\/}. pages 2034--2042.

\bibitem[{Irsoy and Cardie(2014)}]{irsoy-cardie:2014:EMNLP2014}
Ozan Irsoy and Claire Cardie. 2014.
\newblock \href{http://www.aclweb.org/anthology/D14-1080}{Opinion mining with
  deep recurrent neural networks}.
\newblock In {\em Proceedings of the 2014 Conference on Empirical Methods in
  Natural Language Processing (EMNLP)\/}. Association for Computational
  Linguistics, Doha, Qatar, pages 720--728.
\newblock \url{http://www.aclweb.org/anthology/D14-1080}.

\bibitem[{Johansson and Moschitti(2013)}]{Johansson2013RelationalFI}
Richard Johansson and Alessandro Moschitti. 2013.
\newblock {Relational Features in Fine-Grained Opinion Analysis}.
\newblock {\em Computational Linguistics\/} 39:473--509.

\bibitem[{Joty et~al.(2017)Joty, Nakov, M\`{a}rquez, and
  Jaradat}]{joty2017cross}
Shafiq Joty, Preslav Nakov, Llu\'{i}s M\`{a}rquez, and Israa Jaradat. 2017.
\newblock \href{http://aclweb.org/anthology/K17-1024}{{Cross-language Learning
  with Adversarial Neural Networks}}.
\newblock In {\em Proceedings of the 21st Conference on Computational Natural
  Language Learning (CoNLL 2017)\/}. Association for Computational Linguistics,
  Vancouver, Canada, pages 226--237.
\newblock \url{http://aclweb.org/anthology/K17-1024}.

\bibitem[{Jozefowicz et~al.(2015)Jozefowicz, Zaremba, and
  Sutskever}]{jozefowicz2015empirical}
Rafal Jozefowicz, Wojciech Zaremba, and Ilya Sutskever. 2015.
\newblock {An empirical exploration of recurrent network architectures}.
\newblock In {\em Proceedings of the 32nd International Conference on Machine
  Learning (ICML)\/}. pages 2342--2350.

\bibitem[{Katiyar and Cardie(2016)}]{Katiyar2016InvestigatingLF}
Arzoo Katiyar and Claire Cardie. 2016.
\newblock \href{http://www.aclweb.org/anthology/P16-1087}{{Investigating LSTMs
  for Joint Extraction of Opinion Entities and Relations}}.
\newblock In {\em Proceedings of the 54th Annual Meeting of the Association for
  Computational Linguistics (Volume 1: Long Papers)\/}. Berlin, Germany, pages
  919--929.
\newblock \url{http://www.aclweb.org/anthology/P16-1087}.

\bibitem[{Kim and Hovy(2006)}]{Kim2006ExtractingOO}
Soo-Min Kim and Eduard Hovy. 2006.
\newblock \href{http://www.aclweb.org/anthology/W/W06/W06-0301}{{Extracting
  Opinions, Opinion Holders, and Topics Expressed in Online News Media Text}}.
\newblock In {\em Proceedings of the Workshop on Sentiment and Subjectivity in
  Text\/}. Sydney, Australia, pages 1--8.
\newblock \url{http://www.aclweb.org/anthology/W/W06/W06-0301}.

\bibitem[{Kim et~al.(2017)Kim, Stratos, and Kim}]{kim2017adversarial}
Young-Bum Kim, Karl Stratos, and Dongchan Kim. 2017.
\newblock \href{http://aclweb.org/anthology/P17-1119}{Adversarial adaptation of
  synthetic or stale data}.
\newblock In {\em Proceedings of the 55th Annual Meeting of the Association for
  Computational Linguistics (Volume 1: Long Papers)\/}. Association for
  Computational Linguistics, Vancouver, Canada, pages 1297--1307.
\newblock \url{http://aclweb.org/anthology/P17-1119}.

\bibitem[{Kingma and Ba(2015)}]{kingma2014adam}
Diederik Kingma and Jimmy Ba. 2015.
\newblock {Adam: A method for stochastic optimization}.
\newblock In {\em Proceedings of the International Conference on Learning
  Representations (ICLR)\/}. San Diego.

\bibitem[{Li et~al.(2017)Li, Monroe, Shi, Jean, Ritter, and
  Jurafsky}]{li2017adversarial}
Jiwei Li, Will Monroe, Tianlin Shi, S\'ebastien Jean, Alan Ritter, and Dan
  Jurafsky. 2017.
\newblock \href{https://www.aclweb.org/anthology/D17-1229}{Adversarial learning
  for neural dialogue generation}.
\newblock In {\em Proceedings of the 2017 Conference on Empirical Methods in
  Natural Language Processing\/}. Association for Computational Linguistics,
  Copenhagen, Denmark, pages 2147--2159.
\newblock \url{https://www.aclweb.org/anthology/D17-1229}.

\bibitem[{Liu et~al.(2015)Liu, Joty, and Meng}]{liu-joty-meng:2015:EMNLP}
Pengfei Liu, Shafiq Joty, and Helen Meng. 2015.
\newblock \href{http://aclweb.org/anthology/D15-1168}{Fine-grained opinion
  mining with recurrent neural networks and word embeddings}.
\newblock In {\em Proceedings of the 2015 Conference on Empirical Methods in
  Natural Language Processing\/}. Association for Computational Linguistics,
  Lisbon, Portugal, pages 1433--1443.
\newblock \url{http://aclweb.org/anthology/D15-1168}.

\bibitem[{Liu et~al.(2017)Liu, Qiu, and Huang}]{liu2017adversarial}
Pengfei Liu, Xipeng Qiu, and Xuanjing Huang. 2017.
\newblock \href{http://aclweb.org/anthology/P17-1001}{Adversarial multi-task
  learning for text classification}.
\newblock In {\em Proceedings of the 55th Annual Meeting of the Association for
  Computational Linguistics (Volume 1: Long Papers)\/}. Association for
  Computational Linguistics, Vancouver, Canada, pages 1--10.
\newblock \url{http://aclweb.org/anthology/P17-1001}.

\bibitem[{Manning et~al.(2014)Manning, Surdeanu, Bauer, Finkel, Bethard, and
  McClosky}]{manning-EtAl:2014:P14-5}
Christopher~D. Manning, Mihai Surdeanu, John Bauer, Jenny Finkel, Steven~J.
  Bethard, and David McClosky. 2014.
\newblock \href{http://www.aclweb.org/anthology/P/P14/P14-5010}{The {Stanford}
  {CoreNLP} natural language processing toolkit}.
\newblock In {\em Association for Computational Linguistics (ACL) System
  Demonstrations\/}. pages 55--60.
\newblock \url{http://www.aclweb.org/anthology/P/P14/P14-5010}.

\bibitem[{Marcheggiani and Titov(2017)}]{marcheggiani-titov:2017:EMNLP2017}
Diego Marcheggiani and Ivan Titov. 2017.
\newblock \href{https://www.aclweb.org/anthology/D17-1159}{Encoding sentences
  with graph convolutional networks for semantic role labeling}.
\newblock In {\em Proceedings of the 2017 Conference on Empirical Methods in
  Natural Language Processing\/}. Association for Computational Linguistics,
  Copenhagen, Denmark, pages 1506--1515.
\newblock \url{https://www.aclweb.org/anthology/D17-1159}.

\bibitem[{Mart\'{i}nez~Alonso and
  Plank(2017)}]{martinezalonso-plank:2017:EACLlong}
H\'{e}ctor Mart\'{i}nez~Alonso and Barbara Plank. 2017.
\newblock \href{http://www.aclweb.org/anthology/E17-1005}{When is multitask
  learning effective? semantic sequence prediction under varying data
  conditions}.
\newblock In {\em Proceedings of the 15th Conference of the European Chapter of
  the Association for Computational Linguistics: Volume 1, Long Papers\/}.
  Association for Computational Linguistics, Valencia, Spain, pages 44--53.
\newblock \url{http://www.aclweb.org/anthology/E17-1005}.

\bibitem[{Massey~Jr(1951)}]{massey1951kolmogorov}
Frank~J. Massey~Jr. 1951.
\newblock {The Kolmogorov-Smirnov test for goodness of fit}.
\newblock {\em Journal of the American statistical Association\/}
  46(253):68--78.

\bibitem[{Palmer et~al.(2005)Palmer, Kingsbury, and Gildea}]{Palmer2005ThePB}
Martha Palmer, Paul Kingsbury, and Daniel Gildea. 2005.
\newblock {The Proposition Bank: An Annotated Corpus of Semantic Roles}.
\newblock {\em Computational Linguistics\/} 31:71--106.

\bibitem[{Pascanu et~al.(2013)Pascanu, Mikolov, and
  Bengio}]{pascanu2013difficulty}
Razvan Pascanu, Tomas Mikolov, and Yoshua Bengio. 2013.
\newblock {On the difficulty of training recurrent neural networks}.
\newblock In {\em Proceedings of the 31st International Conference on Machine
  Learning (ICML)\/}. pages 1310--1318.

\bibitem[{Pennington et~al.(2014)Pennington, Socher, and
  Manning}]{pennington2014glove}
Jeffrey Pennington, Richard Socher, and Christopher Manning. 2014.
\newblock \href{http://www.aclweb.org/anthology/D14-1162}{{Glove: Global
  Vectors for Word Representation}}.
\newblock In {\em Proceedings of the 2014 Conference on Empirical Methods in
  Natural Language Processing (EMNLP)\/}. Association for Computational
  Linguistics, Doha, Qatar, pages 1532--1543.
\newblock \url{http://www.aclweb.org/anthology/D14-1162}.

\bibitem[{Qin et~al.(2017)Qin, Zhang, Zhao, Hu, and Xing}]{qin2017adversarial}
Lianhui Qin, Zhisong Zhang, Hai Zhao, Zhiting Hu, and Eric Xing. 2017.
\newblock \href{http://aclweb.org/anthology/P17-1093}{Adversarial
  connective-exploiting networks for implicit discourse relation
  classification}.
\newblock In {\em Proceedings of the 55th Annual Meeting of the Association for
  Computational Linguistics (Volume 1: Long Papers)\/}. Association for
  Computational Linguistics, Vancouver, Canada, pages 1006--1017.
\newblock \url{http://aclweb.org/anthology/P17-1093}.

\bibitem[{Reimers and Gurevych(2017)}]{TUD-CS-2017-0196}
Nils Reimers and Iryna Gurevych. 2017.
\newblock Optimal hyperparameters for deep lstm-networks for sequence labeling
  tasks.
\newblock {\em arXiv preprint arXiv:1707.06799\/} .

\bibitem[{Ruppenhofer and Brandes(2016)}]{RUPPENHOFER16.832}
Josef Ruppenhofer and Jasper Brandes. 2016.
\newblock Effect functors for opinion inference.
\newblock In {\em Proceedings of the Tenth International Conference on Language
  Resources and Evaluation (LREC)\/}.

\bibitem[{Ruppenhofer et~al.(2008)Ruppenhofer, Somasundaran, and
  Wiebe}]{Ruppenhofer2008FindingTS}
Josef Ruppenhofer, Swapna Somasundaran, and Janyce Wiebe. 2008.
\newblock {Finding the Sources and Targets of Subjective Expressions}.
\newblock In {\em Proceedings of the Sixth International Conference on Language
  Resources and Evaluation (LREC)\/}. Marrakech, Morocco, pages 2781--2788.

\bibitem[{S{\o}gaard and Goldberg(2016)}]{sogaard2016deep}
Anders S{\o}gaard and Yoav Goldberg. 2016.
\newblock \href{http://anthology.aclweb.org/P16-2038}{Deep multi-task learning
  with low level tasks supervised at lower layers}.
\newblock In {\em Proceedings of the 54th Annual Meeting of the Association for
  Computational Linguistics (Volume 2: Short Papers)\/}. Association for
  Computational Linguistics, Berlin, Germany, pages 231--235.
\newblock \url{http://anthology.aclweb.org/P16-2038}.

\bibitem[{Srivastava et~al.(2014)Srivastava, Hinton, Krizhevsky, Sutskever, and
  Salakhutdinov}]{srivastava2014dropout}
Nitish Srivastava, Geoffrey~E Hinton, Alex Krizhevsky, Ilya Sutskever, and
  Ruslan Salakhutdinov. 2014.
\newblock {Dropout: a simple way to prevent neural networks from overfitting}.
\newblock {\em Journal of machine learning research\/} 15(1):1929--1958.

\bibitem[{Wiebe et~al.(2005)Wiebe, Wilson, and Cardie}]{Wiebe2005AnnotatingEO}
Janyce Wiebe, Theresa Wilson, and Claire Cardie. 2005.
\newblock {Annotating Expressions of Opinions and Emotions in Language}.
\newblock {\em Language Resources and Evaluation\/} 39:165--210.

\bibitem[{Wiegand et~al.(2016{\natexlab{a}})Wiegand, Bocionek, and
  Ruppenhofer}]{wiegand-bocionek-ruppenhofer:2016:N16-1}
Michael Wiegand, Christine Bocionek, and Josef Ruppenhofer. 2016{\natexlab{a}}.
\newblock \href{http://www.aclweb.org/anthology/N16-1094}{Opinion holder and
  target extraction on opinion compounds -- a linguistic approach}.
\newblock In {\em Proceedings of the 2016 Conference of the North American
  Chapter of the Association for Computational Linguistics: Human Language
  Technologies\/}. Association for Computational Linguistics, San Diego,
  California, pages 800--810.
\newblock \url{http://www.aclweb.org/anthology/N16-1094}.

\bibitem[{Wiegand and Ruppenhofer(2015)}]{wiegand-ruppenhofer:2015:CoNLL}
Michael Wiegand and Josef Ruppenhofer. 2015.
\newblock \href{http://www.aclweb.org/anthology/K15-1022}{Opinion holder and
  target extraction based on the induction of verbal categories}.
\newblock In {\em Proceedings of the Nineteenth Conference on Computational
  Natural Language Learning\/}. Association for Computational Linguistics,
  Beijing, China, pages 215--225.
\newblock \url{http://www.aclweb.org/anthology/K15-1022}.

\bibitem[{Wiegand et~al.(2016{\natexlab{b}})Wiegand, Schulder, and
  Ruppenhofer}]{wiegand-schulder-ruppenhofer:2016:N16-1}
Michael Wiegand, Marc Schulder, and Josef Ruppenhofer. 2016{\natexlab{b}}.
\newblock \href{http://www.aclweb.org/anthology/N16-1092}{Separating actor-view
  from speaker-view opinion expressions using linguistic features}.
\newblock In {\em Proceedings of the 2016 Conference of the North American
  Chapter of the Association for Computational Linguistics: Human Language
  Technologies\/}. Association for Computational Linguistics, San Diego,
  California, pages 778--788.
\newblock \url{http://www.aclweb.org/anthology/N16-1092}.

\bibitem[{Wilson(2008)}]{wilson2008fine}
Theresa~Ann Wilson. 2008.
\newblock {\em Fine-Grained Subjectivity And Sentiment Analysis: Recognizing
  The Intensity, Polarity, And Attitudes Of Private States\/}.
\newblock Ph.D. thesis, University of Pittsburgh.

\bibitem[{Wu et~al.(2017)Wu, Bamman, and Russell}]{wu2017adversarial}
Yi~Wu, David Bamman, and Stuart Russell. 2017.
\newblock \href{https://www.aclweb.org/anthology/D17-1187}{Adversarial training
  for relation extraction}.
\newblock In {\em Proceedings of the 2017 Conference on Empirical Methods in
  Natural Language Processing\/}. Association for Computational Linguistics,
  Copenhagen, Denmark, pages 1779--1784.
\newblock \url{https://www.aclweb.org/anthology/D17-1187}.

\bibitem[{Yang and Cardie(2013)}]{Yang2013JointIF}
Bishan Yang and Claire Cardie. 2013.
\newblock \href{http://www.aclweb.org/anthology/P13-1161}{{Joint Inference for
  Fine-grained Opinion Extraction}}.
\newblock In {\em Proceedings of the 51st Annual Meeting of the Association for
  Computational Linguistics (Volume 1: Long Papers)\/}. Sofia, Bulgaria, pages
  1640--1649.
\newblock \url{http://www.aclweb.org/anthology/P13-1161}.

\bibitem[{Yang and Cardie(2014)}]{yang2014joint}
Bishan Yang and Claire Cardie. 2014.
\newblock Joint modeling of opinion expression extraction and attribute
  classification.
\newblock {\em Transactions of the Association for Computational Linguistics\/}
  2:505--516.

\bibitem[{Yang and Mitchell(2017)}]{yang-mitchell:2017:EMNLP2017}
Bishan Yang and Tom Mitchell. 2017.
\newblock \href{https://www.aclweb.org/anthology/D17-1129}{A joint sequential
  and relational model for frame-semantic parsing}.
\newblock In {\em Proceedings of the 2017 Conference on Empirical Methods in
  Natural Language Processing\/}. Copenhagen, Denmark, pages 1258--1267.
\newblock \url{https://www.aclweb.org/anthology/D17-1129}.

\bibitem[{Zhang et~al.(2017)Zhang, Liu, Luan, and Sun}]{zhang2017adversarial}
Meng Zhang, Yang Liu, Huanbo Luan, and Maosong Sun. 2017.
\newblock \href{http://aclweb.org/anthology/P17-1179}{Adversarial training for
  unsupervised bilingual lexicon induction}.
\newblock In {\em Proceedings of the 55th Annual Meeting of the Association for
  Computational Linguistics (Volume 1: Long Papers)\/}. Association for
  Computational Linguistics, Vancouver, Canada, pages 1959--1970.
\newblock \url{http://aclweb.org/anthology/P17-1179}.

\bibitem[{Zhou and Xu(2015)}]{zhou-xu:2015:ACL-IJCNLP}
Jie Zhou and Wei Xu. 2015.
\newblock \href{http://www.aclweb.org/anthology/P15-1109}{End-to-end learning
  of semantic role labeling using recurrent neural networks}.
\newblock In {\em Proceedings of the 53rd Annual Meeting of the Association for
  Computational Linguistics and the 7th International Joint Conference on
  Natural Language Processing (Volume 1: Long Papers)\/}. Beijing, China, pages
  1127--1137.
\newblock \url{http://www.aclweb.org/anthology/P15-1109}.

\end{thebibliography}
